%% file: arxiv.tex
\documentclass{article}

\usepackage{PRIMEarxiv}
\usepackage{cite}
\usepackage{amsmath,amssymb,amsfonts}
\usepackage{stmaryrd}
\usepackage{graphicx}
\usepackage{url}            
\usepackage{fancyhdr}       
\usepackage{algorithm}
\usepackage{algpseudocode}
\usepackage{subfigure}
\usepackage{adjustbox}
\usepackage{multirow}
\usepackage{makecell}
\usepackage{textcomp}
\usepackage{balance}

\begin{document}
\title{Blind Adaptive Local Denoising for CEST Imaging}
\author{Chu Chen\thanks{Corresponding Author (chuchen4-c@my.cityu.edu.hk).} \\
    Department of Mathematics \\ City University of Hong Kong\\
    Hong Kong Centre for \\ Cerebro-cardiovascular Health Engineering\\ Hong Kong \\
\And Aitor Artola\\
    Hong Kong Centre for \\ Cerebro-cardiovascular Health Engineering\\ Hong Kong \\
\And Yang Liu \\ 
    Department of Biomedical Engineering \\City University of Hong Kong \\
    Hong Kong Centre for \\ Cerebro-cardiovascular Health Engineering\\ Hong Kong \\
\And Se Weon Park \\ 
    Hong Kong Centre for \\ Cerebro-cardiovascular Health Engineering\\ Hong Kong \\
\And Raymond H. Chan, \\
    Department of Operations and Risk Management \\ School of Data Science, Lingnan University\\ Hong Kong Centre for \\ Cerebro-cardiovascular Health Engineering\\ Hong Kong \\
\And Jean-Michel Morel \\
    Division of Industrial Data Science \\ School of Data Science, Lingnan University \\ Hong Kong \\
\And Kannie W. Y. Chan \\
    Department of Biomedical Engineering \\City University of Hong Kong \\
    Hong Kong Centre for \\ Cerebro-cardiovascular Health Engineering\\ Hong Kong \\
    Russell H. Morgan Department of Radiology and Radiological Science\\ Johns Hopkins University School of Medicine \\ Baltimore, MD, USA
}
\thanks{This work is partially supported by HKRGC Grants No. CityU11301120, C1013-21GF, CityU11309922, RGC-GRF project 11309925, ITF Grant No. MHP/054/22, LU BGR 105824, and the InnoHK initiative of the Innovation and Technology Commission of the Hong Kong Special Administrative Region Government.}

\maketitle
\input{Abstract}
\input{Introduction}
\input{Related}
\input{Method}
\input{Experiment}
\input{Conlusion}


\balance
\small
\bibliographystyle{unsrt}  
\bibliography{references}  

\end{document}

%% file: Abstract.tex
\begin{abstract}
Chemical Exchange Saturation Transfer (CEST) MRI enables molecular-level visualization of low-concentration metabolites by leveraging proton exchange dynamics. However, its clinical translation is hindered by inherent challenges: spatially varying noise arising from hardware limitations, and complex imaging protocols introduce heteroscedasticity in CEST data, perturbing the accuracy of quantitative contrast mapping such as amide proton transfer (APT) imaging. Traditional denoising methods are not designed for this complex noise and often alter the underlying information that is critical for biomedical analysis. To overcome these limitations, we propose a new Blind Adaptive Local Denoising (BALD) method. BALD exploits the self-similar nature of CEST data to derive an adaptive variance-stabilizing transform that equalizes the noise distributions across CEST pixels without prior knowledge of noise characteristics. Then, BALD performs two-stage denoising on a linear transformation of data to disentangle molecular signals from noise. A local SVD decomposition is used as a linear transform to prevent spatial and spectral denoising artifacts. We conducted extensive validation experiments on multiple phantoms and \textit{in vivo} CEST scans. In these experiments,  BALD consistently outperformed state-of-the-art CEST denoisers in both denoising metrics and downstream tasks such as molecular concentration maps estimation and cancer detection. 
\end{abstract}


\keywords{Chemical Exchange Saturation Transfer, MRI, Denoising, Singular Value Decomposition}

%% file: Introduction.tex
\section{Introduction}
Chemical Exchange Saturation Transfer (CEST) MRI is an advanced imaging technique that enhances the detection of specific molecules within tissues by exploiting the chemical exchange properties of labile protons. CEST MRI works by selectively saturating the proton signals of these molecules, which then transfer the saturation effect to the surrounding water protons through chemical exchange. This transfer leads to a reduction in the water signal that can be detected and quantified, allowing indirect imaging of the molecules of interest. This imaging approach is particularly valuable for identifying low-concentration solutes, such as metabolites, peptides, and proteins~\cite{zhou2011differentiation,haris2014technique,chen2020vivo}, which are otherwise challenging to detect using conventional MRI methods. Clinical interest in CEST imaging is growing due to its ability to provide molecular-level information that can help in the evaluation of various pathologies, including neurological conditions and metabolic studies~\cite{jones2018clinical,jabehdar2023chemical,consolino2020non}.  The sensitivity of the technique to various molecular environments makes it useful for examining changes in tissue composition and pH~\cite{anemone2019imaging}, as well as the presence of biomarkers associated with disease processes~\cite{cember2023glutamate}.

However, achieving reliable CEST measurements can be challenging due to the inherently low signal-to-noise ratio (SNR) associated with the data acquisition process~\cite{jones2013nuclear, zaiss2018chemical}. The CEST signals related to the exchange of labile protons are weak and often overshadowed by system noise, making it difficult to discern meaningful molecular information. This limitation stresses the importance of developing CEST-specific denoising methods to improve the reliability of physical quantities extracted from CEST data.

In fact, the problem of CEST denoising remains a significant area of research because of a number of intrinsic and practical factors. The CEST data are multi-dimensional, encompassing varying frequency offsets and noise characteristics. These variations across different scanners, settings, and subjects require denoising methods to be adaptive. The performance of conventional filters depends on hyperparameter setup, whereas pretrained deep learning models struggle with generalization capability, making them difficult to adjust to the variations automatically. The imperfection of denoising (e.g., over-smoothing or under-processing) can lead to either the loss of crucial diagnostic information or noise artifacts that affect the quantification of CEST effects.

To address these issues, we propose a new algorithm for improving the sensitivity of CEST imaging called Blind Adaptive Local Denoising (BALD). BALD is a two-stage process designed for adaptive denoising of CEST data with unknown noise models and data distributions. BALD starts by estimating the noise model from the given CEST acquisition and then performs an adaptive variance-stabilizing transform (AVST) to ensure spatial homoscedasticity of the noise. BALD then applies patch-wise singular value decomposition (SVD) based denoising to the transformed image, followed by the aggregation of the processed patches and inverse AVST.

Evaluations of the BALD algorithm, along with other baselines, are performed on both phantom and \textit{in vivo} datasets. Quantitative and qualitative results demonstrate that BALD outperforms other methods in terms of CEST image quality and the sensitivity of molecular contrasts.

The main contributions are summarized as follows:
\begin{itemize}
    \item We present the Blind Adaptive Local Denoising (BALD) framework, a CEST-specific denoiser that combines noise homogenization with local structure-spectrum adaptive denoising.
    \item We introduce a novel two-stage algorithm that combines hard thresholding and Wiener filtering in the local SVD decomposition of the data, enabling edge-preserving denoising while adaptively suppressing noise in both spatial and spectral dimensions.
    \item We demonstrate that the proposed variance stabilization transform, AVST, along with its inverse, iAVST, allow neural networks, pre-trained for natural image denoising, to be used for CEST denoising.
    \item Comprehensive evaluations on both phantom and \textit{in-vivo} datasets show the superior performance of BALD over other denoising methods, and show that BALD behaves the best as a pre-processing algorithm for the downstream CEST analysis.
\end{itemize}

%% file: Related.tex
\section{Related Works}
As an essential step in CEST data processing, denoising methods have received increasing attention to obtain more accurate molecular information. Due to the unique structure of CEST data, denoising approaches relying solely on images or CEST signals (z-spectrum) often perform poorly~\cite{chen2020high,romdhane2021evaluation,chen2024implicit}. Effective CEST denoising requires the use of spatial-temporal correlation to maximize denoising performance. The multilinear singular value decomposition (MLSVD)~\cite{chen2020high} exploits this correlation and the low-rank property of the CEST tensor by truncation of the decomposed core tensor. However, direct truncation of the coefficients projected by SVD can hardly fully remove noise-related components while well preserving signal-related ones. Romdhane \textit{et al.}~\cite{romdhane2021evaluation} combined Non-local means and coherence-enhancing diffusion (NLmCED) to improve CEST imaging. NLmCED tends to create artifacts when the data is heavily corrupted with low SNR, and the gradient computation fails to capture signal variations~\cite{chen2024implicit}. BOOST~\cite{chen2024boosting} leverages variance-stabilizing transformation (VST) and spatial-spectral redundancy through tensor subspace decomposition, where noise components are identified using a low-rank approximation and spectral smoothness constraints. However, the predefined Rician-distributed noise model limits the performance of VST and filtering, and the tradeoff regularization parameters require empirical tuning.

Recent advances reveal that deep learning frameworks improve the sensitivity of CEST imaging. DECENT~\cite{chen2023learned} introduces a global-spectral convolution neural network that explicitly learns spatiotemporal correlation priors, while DCAE-CEST~\cite{kurmi2024enhancing} employs an autoencoder trained on simulated z-spectra with Kullback-Leibler divergence constraints, and validated via adaptive learning using principal component analysis. Effective to a certain extent, though, the performance of pre-trained models degrades when processing data with a varying acquisition protocol. A paradigm shift is observed in IRIS~\cite{chen2024implicit}, which formulates denoising as a regression problem by projecting CEST data onto an orthogonal subspace with an implicit neural representation. IRIS achieves a satisfying sensitivity while maintaining computational efficiency. However, global truncated SVD is unable to precisely handle the anisotropic noise distribution presented in CEST data.

To address the limitations of the aforementioned methods, we introduce the rationale and details about the proposed BALD algorithm in the following sections.

%% file: Method.tex
\section{Methods}

\begin{figure*}[!t]
\centering
  \includegraphics[width=1\linewidth]{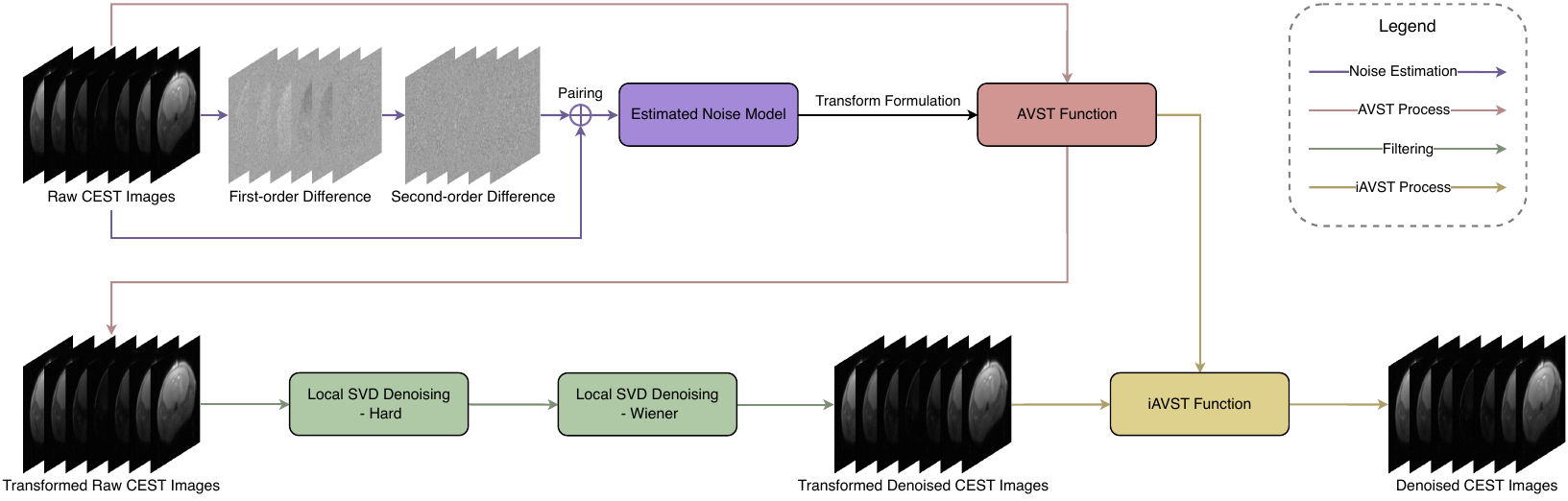}
  \caption{Schematic illustration of the Blind Adaptive Local Denoising (BALD) Framework.}
  \label{flowchart}
\end{figure*}

\subsection{Adaptive Variance Stabilization}
\label{sec:avst}
According to our analysis (refer to Section~\ref{sec:noise_modeling}) and existing works~\cite{zaiss2011quantitative,chen2020high,chen2024boosting}, the variance of noise introduced by the CEST acquisition protocol is signal-dependent. We propose an Adaptive Variance Stabilization Transform (AVST), tailored to the noise characteristics of each specific CEST acquisition. We consider the following signal-dependent additive noise model of a CEST sequence $Y$ of $C$ frequencies:
\begin{equation}
	\tilde{u}_i(\mathbf{x}) = u_i(\mathbf{x}) + g(u_i(\mathbf{x}))n_i(\mathbf{x}), \quad \text{with} \quad n_i(\mathbf{x}) \sim \mathcal{N}(0, 1),
    \label{eq:noise}
\end{equation}
where $\tilde{u}_i(\mathbf{x})$ and $u_i(\mathbf{x})$ are respectively the noisy and noiseless pixels at the position $\mathbf{x}\in\Omega$ and the offset frequency $i\in \llbracket 1,C\rrbracket$, $g(u_i)$ is the noise model that quantifies the signal-dependent noise deviation (also called noise curve), and $\Omega$ is the set of all possible spatial coordinates in the sequence.

Let $f$ be a variance-stabilizing transformation. Using a first-order Taylor expansion of $f(\tilde{u}_i)$ around $u_i$, we obtain
\begin{equation}
	f(\tilde{u}_i(\mathbf{x})) \simeq f(u_i(\mathbf{x})) + f'(u_i(\mathbf{x}))g(u_i(\mathbf{x}))n_i(\mathbf{x}).
\end{equation}
To make the noise signal-independent, it is enough to choose the transform $f$ such that $f'(u)g(u) = \sigma$ where $\sigma>0$ is the standard deviation of the targeted fixed noise. Therefore, $f$ is defined as the integral of the inverse of the noise model $g$ multiplied by $\sigma$,
\begin{equation}
	f(u) = \sigma \int_{0}^{u} \frac{1}{g(t)}dt.
    \label{eq:transform}
\end{equation}

As explored by previous works~\cite{kim2015review,woessner2005numerical}, the z-spectra are described by physical models that are smooth and second-order differentiable with respect to the saturation frequency. To build an empirical noise model, AVST assumes that the signal is nearly affine and estimates $g(u_i)$ by applying normalized second-order finite differences along the saturation frequency dimension, which suppresses the signal and leaves only the noise. Consider a z-spectrum at any position $\mathbf{x}$, let $\tilde{u}_k$ and $u_k$ denote, respectively, the noisy and noise-free intensities, at the $k$-th ($1<k<C$) saturation frequency offset.
\begin{align}
    \Delta u_k & = 2u_k - u_{k-1} - u_{k+1} \approx 0\\
    \Delta \tilde{u}_k& \approx 2g(u_k) n_k - g(u_{k-1}) n_{k-1} - g(u_{k+1})  n_{k+1}
\end{align}
Since the sum of Gaussian random variables is Gaussian, the second derivative of the signal along $k$ is also Gaussian, with the following parameters,
\begin{align}
    \mathbb{E}[\Delta \tilde{u}_k] & = 0 \\ \nonumber
    Var(\Delta \tilde{u}_k) & =  4g(u_k)^2 + g(u_{k-1})^2 + g(u_{k+1})^2\\
    & \approx 6g(u_k)^2\\
    \Delta \tilde{u}_k & \sim \mathcal{N}(0,6g(u_k)^2).
\end{align}
The last approximation also arises from the signal's smoothness along $k$. Thus, the local noise is estimated as 
\begin{equation}
	\bar{n}(\tilde{u}_k) = \frac{1}{\sqrt{6}}\Delta \tilde{u}_k\sim \mathcal{N}(0,g(u_k)^2).
        \label{eq:noise_extract}
\end{equation}
This formulation exploits the smoothness of CEST spectra, where abrupt signal changes arise primarily from noise rather than actual biological contrast. The use of saturation-frequency differences explicitly models noise correlations along the spectroscopic dimension, unlike traditional intensity-only transforms such as Anscombe VST. Noise modeling in AVST applies second-order differences over first-order differences based on the fact that noise fluctuation in CEST data arises from stochastic acquisition errors, which manifest as high-frequency perturbations. First-order differences conflate these perturbations with directional intensity gradients (e.g., ascending/descending slopes in Z-spectra), leading to overestimation of $g(u)$. Second-order differences, by focusing on local curvature, inherently filter out linear signal trends and extract better the local noise.
Using Eq.~(\ref{eq:noise_extract}), we extracted the noise for all  frames except the first and last. For each pixel, we have an intensity-noise pair $(\tilde{u}_i(\mathbf{x}), \bar{n}(\tilde{u}_i(\mathbf{x})))$ that we will use to compute the noise curve. Since the CEST acquisition is performed sequentially along $k$, like a video, some frames may be corrupted by subject motion, which will corrupt the noise extracted by the second derivative. To obtain a robust estimate, we first compute noise curves independently for each frame and then merge them using the median to reduce the impact of outliers.

The noise curve for each frame is computed by grouping its pairs by intensity. The intensity dimension is divided into $t_1\in \mathbb{Z}^+$ equally spaced bins. The $l$-th bin $b^k_l$ of the $k$-th frame,

\begin{equation}
    b_k^l = \left\{\bar{n}(\tilde{u}_i(\mathbf{x}))|\frac{(l-1)}{t_1}L<\tilde{u}_k(\mathbf{x})<\frac{l}{t_1}L\right\},
\end{equation}
where the intensity range $L=\underset{\mathbf{x}\in\Omega,i\in\llbracket1,C\rrbracket}{\max}~\tilde{u}_i(\mathbf{x}) - \underset{\mathbf{x}\in\Omega,i\in\llbracket1,C\rrbracket}{\min}~\tilde{u}_i(\mathbf{x})$.

The standard deviation of the noise in each bin at the middle of the bin is used as a first rough estimate of the noise curve,
\begin{align}
    \textbf{g}_l^k &= \left(\frac{l+\frac{1}{2}}{t_1}L, g_l^k\right), \quad \text{with} \quad g_l^k = \sqrt{\frac{1}{|b_l^k|}\sum_{n\in b_l^k}n^2}.
\end{align}
Then, as mentioned, the frame noise curves are merged using the median.
\begin{align}
    \textbf{g}_l = \left(\frac{l+\frac{1}{2}}{t_1}L, \text{median}\left(g_l^2,...,g_l^{C-1}\right)\right)
\end{align}
The continuous noise model $g(\cdot)$ can then be obtained by linear interpolation of $\textbf{g}_l$ to any desired length $t_2$ $(t_2\gg t_1)$.

The initial sparse sampling of $t_1$ intervals guarantees that each bin contains enough pixels to compute reliable statistics, mitigating the risk of overfitting to sparse or noisy local estimates. Since practical CEST acquisitions may suffer from uneven sampling intervals along the saturation frequency dimension, which can introduce spurious high $g(u_i)$ values due to interpolation artifacts or motion-induced misalignment, the selection of the representative median value aims to discard outliers in $g(u_i)$.

In AVST, we chose the targeted noise level $\sigma$ to be the average of the noise curve,
\begin{equation}
\sigma=\frac{1}{t_2} \sum_{i=1}^{t_2} g(u_i).
\label{eq:target_noise}
\end{equation}

The transformation $f(u)$ can then be computed by Eq.~(\ref{eq:transform}) with the constructed $g(u)$, and the AVST stabilizes the variance of the target CEST sequence through $\hat{Y} = f(Y)$.

\subsection{Filtering with Local Orthogonal Projection}
The denoising pipeline is a two-stage linear transform Wiener-type method. The two-stage framework~\cite{lebrun2012secrets} refers to methods where a preliminary ``rough'' denoising is first applied to estimate an oracle of the noiseless image. This oracle is then used as a reference to fine-tune the second denoising stage. 
In essence, our method is inspired by two-stage DCT denoising~\cite{lebrun2012secrets,pierazzo2017multi}. This method performs an initial denoising by hard thresholding of the DCT decomposition. It relies on the linearity of the DCT and the fact that the spectrum of natural images is sparse, while the spectrum of noise is uniform. Then, the first denoised oracle is used to estimate the optimal Wiener coefficient and perform soft thresholding in the DCT space. This approach can be applied with any kind of linear transform, such as the Fourier or wavelets, with good properties. 

Unlike natural images, CEST data have many channels with smooth transitions in the channel dimension, making it more similar to video data. Due to this difference, we can apply a 1D linear transform directly to the depth dimension rather than a 2D spatial transform on 2D patches. Given this and the local similarity of CEST, we decided to use SVD decomposition, computed at the patch level, as a linear transformation of the CEST pixels. The SVD transform is known to produce a sparse representation adapted for denoising~\cite{mairal2007sparse}. The denoising is applied to the variance-stabilized image $\hat{Y}$.

\subsubsection{Hard Thresholding for Initial Denoising} The first stage, implemented via \textsc{SVDDenoisingHard}(Algorithm~\ref{alg:hard}), applies singular value decomposition (SVD) to local image patches to facilitate signal and noise separation. For each tensor patch of size $s\times s \times C$, flattened along the spatial dimension $Y_{patch}\in \mathbb{R}^{s^2\times C}$, the singular value decomposition (SVD) yields
\begin{equation}
    Y_{patch} = U\Sigma V^T
\end{equation}
where $U$ and $V$ are orthogonal bases and $\Sigma$ is the diagonal matrix containing the singular values. The spatial coefficients $\hat{U}=U\Sigma$ are thresholded to suppress noise,
\begin{equation}
\hat{U}(\omega) = \left\{
             \begin{array}{ll}
             0 & \text{if}~ \hat{U}(\omega) < 3\sigma, \\
             \hat{U}(\omega) & \text{otherwise}.
             \end{array}
\right.
\end{equation}
Here, $\sigma$ is the global noise deviation estimated from Eq.~(\ref{eq:target_noise}), and the threshold $3\sigma$ ensures a high probability of retaining true signal components under Gaussian noise assumptions. The adaptive weight $(1+N_p)^{-1}$, where $N_p$ counts retained coefficients, balances contributions from patches with varying sparsity levels during aggregation.

\subsubsection{Wiener Filtering for Signal-Adaptive Refinement} The second stage, \textsc{SVDDenoisingWiener} (Algorithm~\ref{alg:wiener}), utilizes the oracle image $G$ (output from \textsc{SVDDenoisingHard}) to compute optimal shringkage factors. For each patch, the Wiener weight $\rho(\omega)$ is derived as
\begin{equation}
    \rho(\omega) = \frac{U_G(\omega)^2}{U_G(\omega)^2 + \sigma^2}
\end{equation}
where $U_G = GV$ represents the projection of the guide image onto the noise-whitened subspace. This formulation adaptively attenuates components where noise dominates ($U_G\ll \sigma$) while preserving those with strong signal ($U_G\gg \sigma$). The cumulative weight $S_P=\sum \rho(\omega)^2$ quantifies patch-wise signal fidelity, and the aggregation weight $(1 + S_P)^{-1}$ prioritizes patches with higher signal-to-noise ratios (SNR). Under mild independence assumptions, the aggregation weight of both stages and the Wiener coefficients can be proved to provide optimal denoising~\cite{dabov2007image}.

Our method enhances the SVD denoising proposed in MLSVD~\cite{chen2020high} in two ways.
First, by operating on small spatial-temporal patches, the method exploits local structural correlations in CEST sequences while avoiding global over-smoothing. Second, the two-step refinement mechanism leverages the hard thresholding stage to remove bulk noise and give a first denoised estimate that is used as an oracle to perform near-optimal Wiener filtering in the second step. 

\subsection{BALD Algorithm}
\begin{algorithm}
\caption{Blind Adaptive Local Denoising (BALD)}
\label{alg:bald}
\begin{algorithmic}[1]
\Function{BALD}{$Y, t_1, t_2, s$}
  \State \textbf{input:} noisy image $Y$, number of original intervals $t_1$, number of interpolated intervals $t_2$, and patch size $s$.
  \State \textbf{output:} denoised image $X$.
  \State $[\hat{Y}, g] \gets \text{AVST}(Y, t_1, t_2)$

  \State $\sigma \gets \frac{1}{t_2} \sum_{i=1}^{t_2} g_i$
  \State $G \gets \text{SVDDenoisingHard}(\hat{Y}, \sigma, s)$
  \State $\hat{X} \gets \text{SVDDenoisingWiener}(\hat{Y}, G, \sigma, s)$
  \State $X \gets \text{iAVST}(\hat{X},g)$
  \State \textbf{return} $X$
\EndFunction
\end{algorithmic}
\end{algorithm}
Fig.~\ref{flowchart} illustrates the complete workflow of the Blind Adaptive Local Denoising (BALD) algorithm, with pseudocode formally detailed in Algorithm~\ref{alg:bald}. The BALD framework leverages a stabilize-denoise-restore paradigm where AVST and local filtering operate synergistically: 1. The noisy sequence $Y$ first undergoes AVST to estimate the variance-stabilizing transform $f(\cdot)$ from the data and transform the signal-dependent noise into approximately additive Gaussian noise with stabilized variance; 2. The 2-stage local SVD filter processes the transformed sequence $\hat{Y}$ with estimated noise deviation $\sigma$ from upstream; 3. The denoised stabilized sequence $\hat{X}$ is mapped back to the original intensity domain via the inverse AVST (iAVST) $f^{-1}(\cdot)$ and outputs the final denoised CEST sequence $X$.

By converting signal-dependent noise into stationary Gaussian noise, AVST unlocks the full potential of SVD-based local filtering, which assumes additive white Gaussian noise. The patch-wise adaptive orthogonal decomposition in SVD denoising respects local correlations in both spatial and saturation-frequency dimensions, preventing over-smoothing of CEST-specific spectral features stabilized by AVST. This cohesive design ensures robust denoising performance while maintaining the quantitative accuracy essential for CEST imaging, particularly in physiologically critical low-intensity zones where conventional methods falter.

\begin{algorithm}
\caption{Local SVD Denoising - Hard}
\label{alg:hard}
\begin{algorithmic}[1]
\Function{SVDdenoisingHard}{$Y, \sigma, s$}
  \State \textbf{input:} noisy image $Y$, estimated noise deviation $\sigma$, and patch size $s$.
  \State \textbf{output:} denoised image $X$
  \State $X, W \gets 0$
  \For{each patch domain $\Omega_{\text{patch}} \subseteq \Omega$ of size $s \times s$} 
  \State $Y_{patch} \gets \text{flatten}(Y(\Omega_{\text{patch}}))$
      \State $[U, \Sigma, V] \gets \text{SVD}(Y_{patch})$ 
      \State $\hat{U} \gets U\Sigma$ 
      \For{$\omega \in (\{1,\dots,s^2\}\times\{1,\dots,c\})$}
      \If{$\hat{U}(\omega) < 3\sigma$} 
      \State $\hat{U}(\omega) \gets 0$
      \Else 
      \State $N_p \gets N_p + 1$
      \EndIf
      \EndFor
      \State $b_{\text{tmp}} \gets U \Sigma V^T$
    \State $X(\Omega_{\text{patch}}) \gets X(\Omega_{\text{patch}}) + b_{\text{tmp}} \cdot (1 + N_p)^{-1}$
    \State $W(\Omega_{\text{patch}}) \gets W(\Omega_{\text{patch}}) + (1 + N_p)^{-1}$ 
  \EndFor
  \State $X \gets X/W$ 
  \State \textbf{return} $X$
\EndFunction
\end{algorithmic}
\end{algorithm}

\begin{algorithm}
\caption{Local SVD Denoising - Wiener}
\label{alg:wiener}
\begin{algorithmic}[1]
\Function{SVDdenoisingWiener}{$Y, G, \sigma, s$}
  \State \textbf{input:} noisy image $Y$, guide image $G$, estimated noise deviation $\sigma$, and patch size $s$
  \State \textbf{output:} denoised image $X$
  \State $X, W \gets 0$
  \For{each patch domain $\Omega_{\text{patch}} \subseteq \Omega$ of size $s \times s$} 
  \State $S_p \gets 0$
  \State $Y_{patch} \gets \text{flatten}(Y(\Omega_{\text{patch}}))$
      \State $[U, \Sigma, V] \gets \text{SVD}(Y_{patch})$ 
      \State $\hat{U} \gets U\Sigma$
      \State $U_G \gets GV$
      \For{$\omega \in (\{1,\dots,s^2\}\times\{1,\dots,C\})$}
      \State $\rho(\omega) \gets \frac{U_G(\omega)^2}{U_G(\omega)^2 + \sigma^2}$
      \State $S_P \gets S_P + \rho(\omega)^2$
      \State $\hat{U}(\omega) \gets \hat{U}(\omega)\rho(\omega)$
    \EndFor
    \State $b_{\text{tmp}} \gets \hat{U}V$
    \State $X(\Omega_{\text{patch}}) \gets X(\Omega_{\text{patch}}) + b_{\text{tmp}} \cdot (1 + S_P)^{-1}$
    \State $W(\Omega_{\text{patch}}) \gets W(\Omega_{\text{patch}}) + (1 + S_P)^{-1}$ 
  \EndFor
  \State $X \gets X/W$ 
  \State \textbf{return} $X$
\EndFunction
\end{algorithmic}
\end{algorithm}

\subsection{CEST Analysis}
CEST analysis aims to extract quantitative contrast information from z-spectra~\cite{kim2015review}.  Numerous techniques have been proposed for the mapping of CEST effect~\cite{woessner2005numerical,chen2019creatine,glang2020deepcest,huang2022deep,huemer2024improved,chen2025high}. A straightforward and widely used method for generating CEST contrast, particularly for Amide Proton Transfer-weighted (APTw) imaging, is the Magnetization Transfer Ratio Asymmetry ($\text{MTR}_{\text{asym}}$) analysis. The $\text{MTR}_{\text{asym}}$ at a specific offset $\Delta\omega$ is calculated as the difference between the z-spectrum values at the negative and positive offsets, i.e., 
\begin{equation} 
\text{MTR}_{\text{asym}}(\Delta\omega) = z(-\Delta\omega) - z(\Delta\omega). 
\end{equation} 
For APTw imaging, the contrast is obtained by evaluating $\text{MTR}_{\text{asym}}$ at the amide proton frequency offset of 3.5 ppm. Among established techniques, Lorentzian line-shape analysis~\cite{zaiss2011quantitative} offers significant advantages by explicitly modeling saturation transfer effects. Multi-pool Lorentzian Fitting (MPF) extends this approach to characterize multiple exchange pools (e.g., amide, amine, hydroxyl) by solving the nonlinear optimization problem,
\begin{equation}
\hat{p} = \operatorname*{argmin}_{p} \frac{1}{2} || \mathcal{L}(p, \Delta\omega) - z ||_2^2,
\end{equation}
where $p$ represents the parameters related to exchangeable pools, $\Delta\omega$ is saturation frequency (offsets) and $\mathcal{L}$ is the Lorentzian model defined as
\begin{equation}
    \mathcal{L}(p, \Delta\omega) = 1 - \frac{A_{DS}}{1+(\frac{\Delta\omega-\delta_{DS}}{\Gamma_{DS}/2})^2} - \sum_{j\in pools} \frac{A_i}{1+(\frac{\Delta\omega-\delta_{DS}-\delta_{j}}{\Gamma_{j}/2})^2},
    \label{lormodel}
\end{equation}
where direct saturation ($DS$) represents the water pool, and $pools$ is the set of indices of the target molecular pools that are exchanged with water. For an $n$-pool Lorentzian model, each pool is parameterized by its amplitude $A_{j}$ and peak width $\Gamma_{j}$, as well as the center frequency for direct saturation $\delta_{DS}$. Being one of the most commonly used methods, which contributes to significant pathological findings, we adopt MPF in this work for CEST analysis, validating the denoising baselines through structural correlation and pathological characteristics of the extracted CEST contrasts.

%% file: Experiment.tex
\section{Experiments}
In this section, we conduct comprehensive evaluations of BALD across both phantom and \textit{in vivo} datasets, in comparison to other denoising baselines, including BM4D~\cite{maggioni2012nonlocal}, NLmCED~\cite{romdhane2021evaluation}, MLSVD~\cite{chen2020high}, DECENT~\cite{chen2023learned}, BOOST~\cite{chen2024boosting}, and IRIS~\cite{chen2024implicit}. 
\subsubsection{Data Preparation}
The synthetic phantom was generated with Bloch-McConnell-based z-spectrum simulation using CESTsimu~\cite{zhang2025cestsimu}. The phantom was constructed as a 3×3 grid of synthetic CEST samples with orthogonal concentration gradients for key exchangeable pools. APT (amide proton transfer at 3.5 ppm) concentration increased column-wise from 0M (left column) to 0.4M (middle) and 0.8M (right column), while NOE (relayed nuclear Overhauser effect at -3.5 ppm) concentration increased row-wise from 0M (top row) to 0.8M (middle) and 1.6M (bottom row). All phantoms shared a uniform background containing a water pool and a 15M MT (magnetization transfer at -2.5 ppm) pool. Simulations were performed at $B_0$ = 3T field strength using continuous wave saturation with $B_1$ = 0.8 $\mu T$. Z-spectra were sampled with saturation frequency offsets ranging from -10 ppm to 10 ppm with 0.25 ppm increments. The $\textit{M}_0$ image for signal normalization was acquired at 300 ppm offset, with all other Bloch-McConnell parameters (e.g., exchange rates, relaxation times) set to established biological defaults. This design created systematic concentration variations across both dimensions, allowing a comprehensive evaluation of denoising performance under controlled exchange pool conditions.

For the real phantom, Dopamine Hydrochloride was dissolved in phosphate-buffered saline (PBS at pH 5.5) at concentrations of 20 mM and 50 mM as shown in Fig.~\ref{fig:realphan_vis}. The samples were placed at room temperature for 72 hours before scanning. The CEST MRI sequence was a continuous-wave (CW) saturation module followed by rapid acquisition with refocused echoes (RARE) as a readout module, and acquired for three individual scans of the same phantom. A power ($B_1$) of $1.2 \mu T$ and a duration ($t_{sat}$) of 3000 ms were used for the saturation module. The saturation frequency varied from -15 to 15 ppm, with a 0.2 ppm increment between -7 and 7 ppm and a 1 ppm increment from -15 to -6 ppm and from 6 to 15 ppm. Four $\textit{M}_0$ images with saturation frequency offset at 300 ppm were acquired and averaged for z-spectrum normalization. The readout parameters were as follows: repetition time (TR)=5000 ms, echo time (TE)=4.7 ms, matrix size=$64\times64$ within a field of view (FOV) of $30\times30$ mm$^2$, slice thickness=1 mm, RARE factor=32.

Given that CEST contrast exhibits significant pathological features in mouse tumors, we tested the effectiveness and rationale of BALD using \textit{in vivo} datasets. All animal experiments were approved by the Animal Ethics Sub-Committee and followed the institutional guidelines of the Institutional Laboratory Animal Research Unit of the City University of Hong Kong, and were conducted under an authorized animal experiment license. We prepared 10 sets of mouse data. Each of them was injected with U-87 MG cell ($0.5 M$/$3ul$) at 2.0 mm right-lateral, 0.2 mm anterior, and 3.8 mm below the bregma~\cite{park2024preclinical}. The same CEST sequence was set up as in real phantom acquisition, except that $B_1 = 0.8 \mu T$; the saturation frequency varied from -15 to 15 ppm, with a 0.2 ppm increment between -7 and 7 ppm and a 2 ppm increment from -15 to -7 ppm and from 7 to 15 ppm; Four $\textit{M}_0$ images were acquired at the saturation frequency offset of 200 ppm; echo time (TE)=5.9 ms, and matrix size=$96\times96$. Another individual's data set with the same acquisition configuration was scanned for APTw studies, but with $B_1 = 0.8 \mu T$ and matrix size=$64\times64$.

All MRI experiments were performed on a horizontal bore 3T Bruker BioSpec system (Bruker, Ettlingen, Germany).

\subsubsection{Implementation Details}
In all experiments, the number of original intervals $t_1$ and the number of interpolated intervals $t_2$ in BALD were set to 10 and 100, respectively. The patch size $s$ was chosen between 8 and 16 according to the image resolution and geometric complexity. The default settings were used when implementing the denoising baselines. All experiments were conducted in MATLAB (R2023a) or Python 3.8.17 on a PC equipped with Intel\textsuperscript{\textcircled{\textsc{r}}} Xeon\textsuperscript{\textcircled{\textsc{r}}} Silver 4210 Processor CPU 2.20GHz and Nvidia GeForce RTX 3090 GPU with 24G of memory.

\subsection{Synthetic Phantom}

\begin{figure*}
\centering
    \subfigure[Comparison on CEST Sequence]{
{\includegraphics[width=0.25\linewidth]{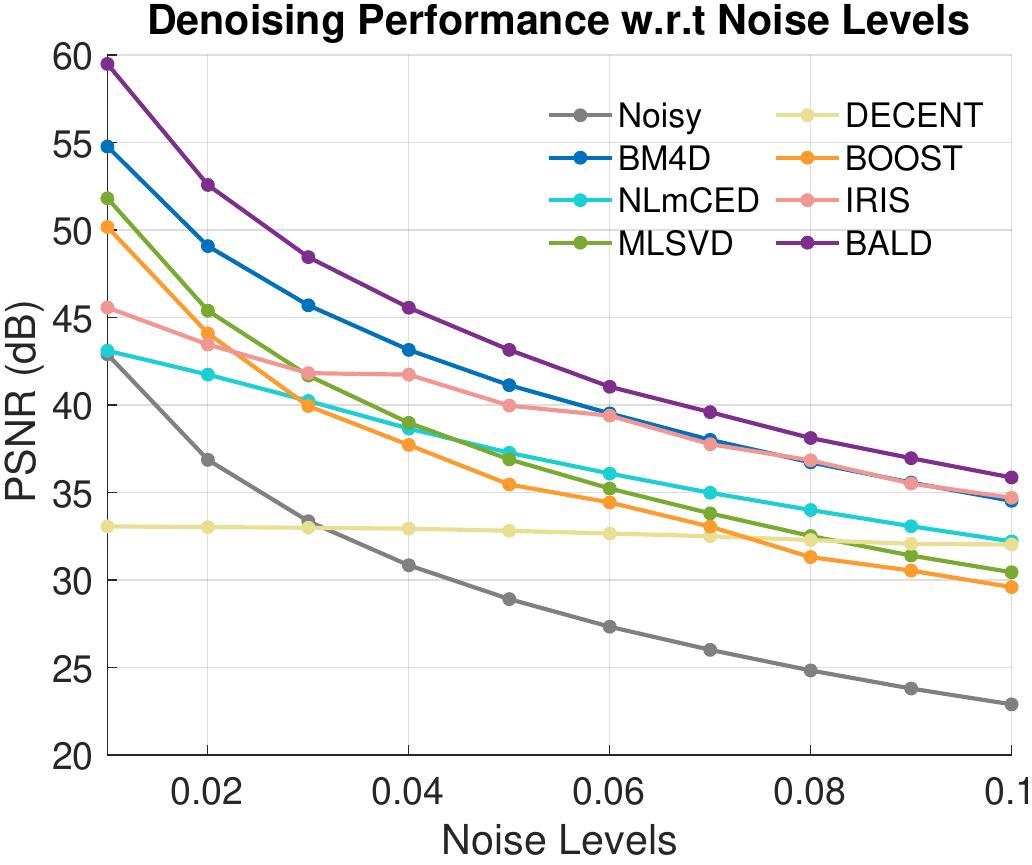}}}
   \subfigure[Comparison on CEST Contrasts]{
{\includegraphics[width=0.4\linewidth]{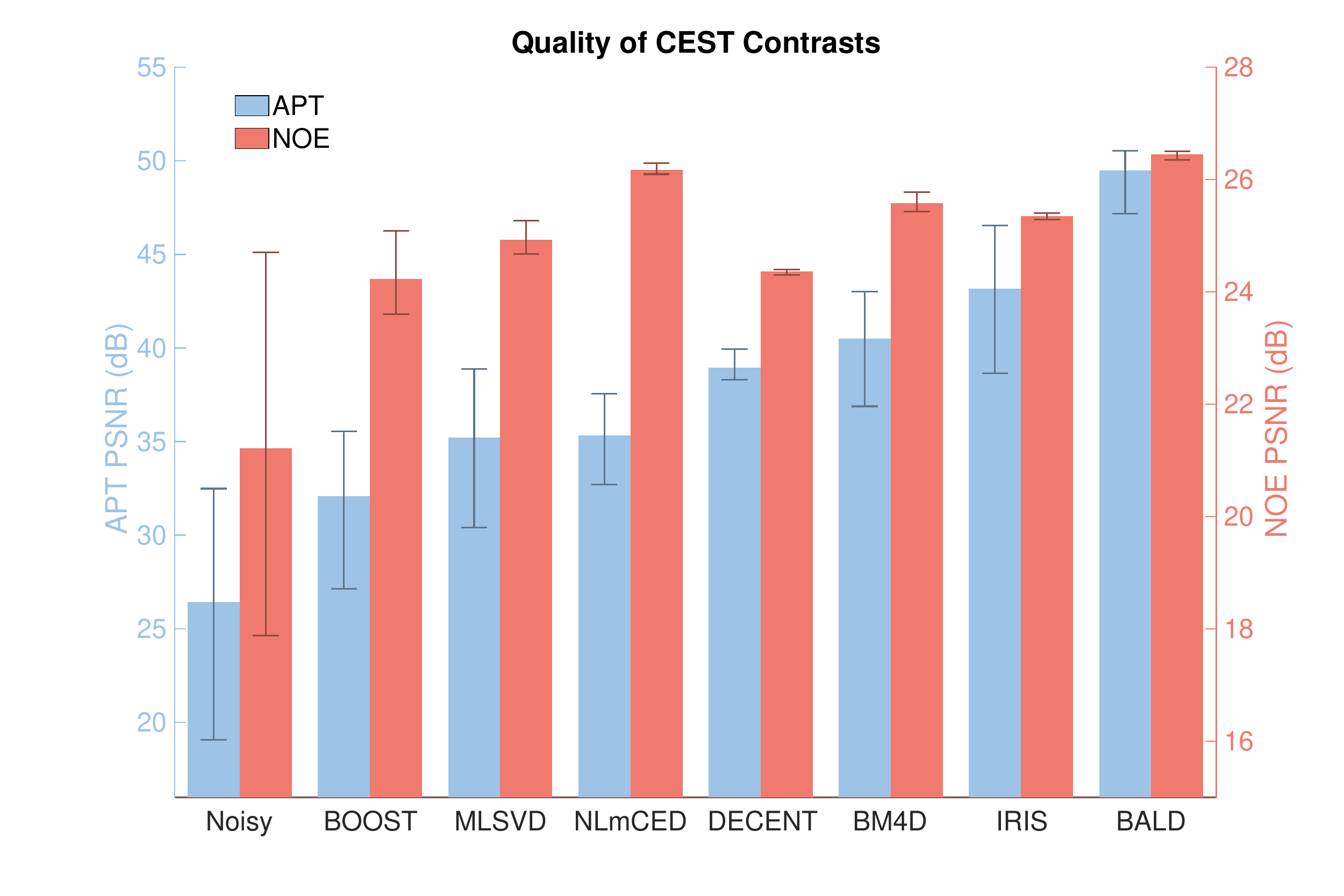}}}
    \subfigure[Reference APT (Left) and NOE (Right) Maps]{
{\includegraphics[width=0.3\linewidth]{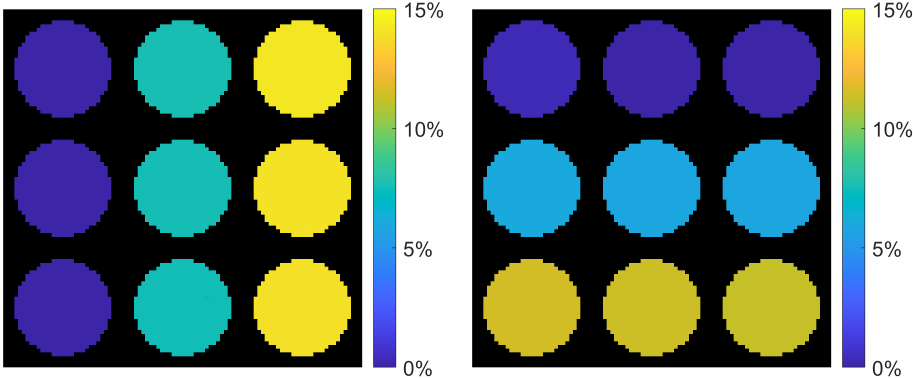}}}
    \subfigure[Visualization of APT (Left) and NOE (Right) Maps]{
{\includegraphics[width=1\linewidth]{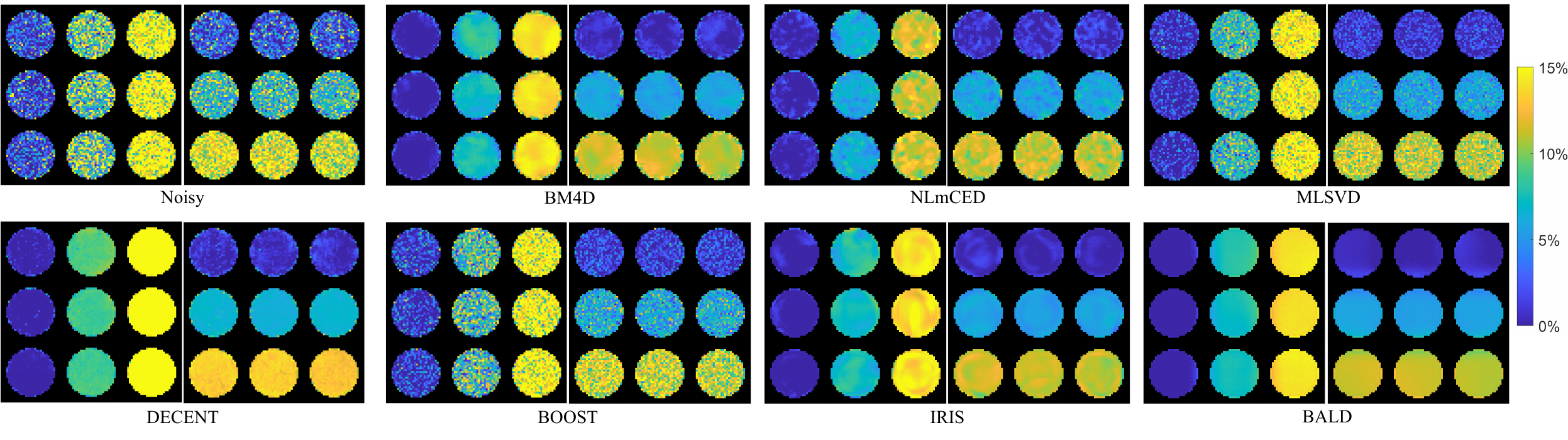}}}
    \caption{Evaluation on synthetic phantoms corrupted by multi-level Rician noise. Quantitative comparisons are based on the (a) CEST sequence and (b) extracted contrasts by multi-pool Lorentzian fitting, respectively. The visualization of CEST contrasts (d) is based on the synthetic phantom under the noise level of 0.05.}
    \label{fig:simphan}
\end{figure*}

The denoising efficacy of BALD was rigorously validated on synthetic phantoms through a two-tier evaluation framework. First, CEST image sequences were corrupted with Rician noise at various levels (0.01–0.1 standard deviation) to assess robustness. As depicted in Fig.~\ref{fig:simphan}(a), BALD consistently outperformed all comparative methods, achieving PSNR gains exceeding 1 dB universally, demonstrating superior noise suppression irrespective of degradation severity. Second, the downstream impact on CEST analysis was evaluated by extracting APT and NOE contrasts via MPF. Fig.~\ref{fig:simphan}(b) reveals that despite MPF's inherent noise sensitivity, BALD-preprocessed data yielded optimally recovered contrasts with the highest fidelity to ground truth (Fig.~\ref{fig:simphan}(c)). The compact interquartile ranges (IQRs) in BALD's contrast PSNR distributions, significantly narrower than alternatives, further attest to its robustness against noise-level variations. Visual comparisons of contrasts at 0.05 std noise (Fig.~\ref{fig:simphan}(d)) provide critical insights: other denoisers exhibited residual outliers and elevated variance within phantoms due to incomplete z-spectrum denoising, while pre-trained model DECENT introduced a systemic contrast shift from distributional mismatch. BALD uniquely balances spatial smoothness with spectral fidelity, eliminating noise without distorting the concentration gradients, validating its adaptability to CEST-specific inverse problems.

\subsection{Real Phantom}

\begin{figure}[ht]
\centering
  \includegraphics[width=1\linewidth]{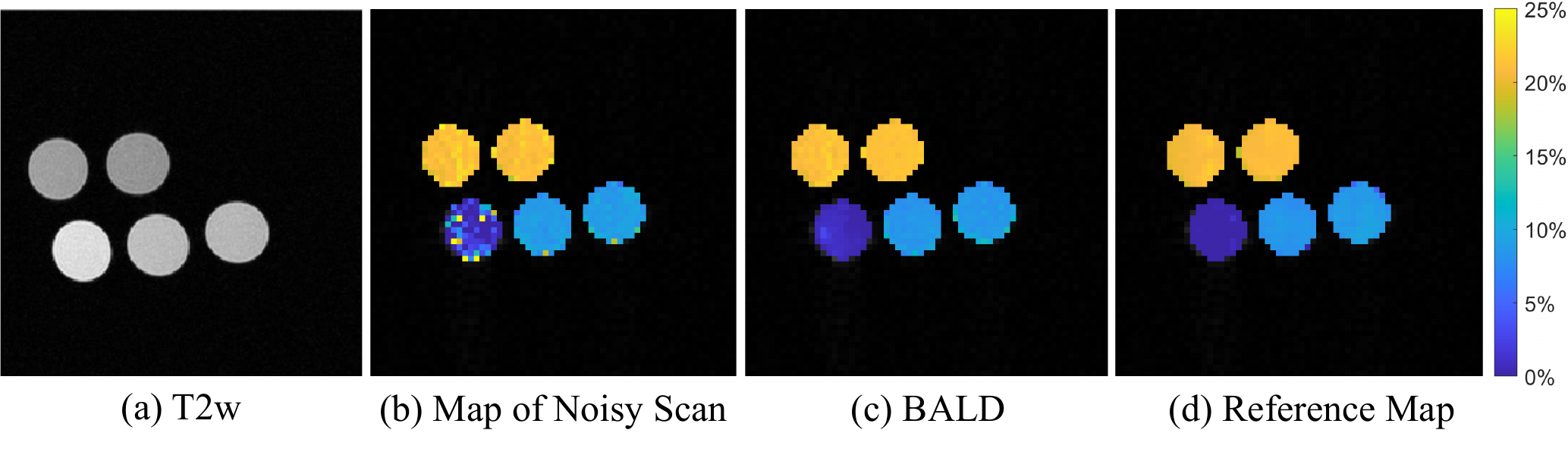}
  \caption{Visualization of dopamine maps generated from BALD processed scan.}
  \label{fig:realphan_vis}
\end{figure}

\begin{table*}[htbp]
\centering
\small
\caption{Quantitative Comparison of Denoising Methods and the Function of AVST on Real Phantom. Each method is tested without/with AVST.}
\begin{tabular}{c|c c c c c c c c c c c c c c}
\hline
        Scan & BM4D~\cite{maggioni2012nonlocal} &  NLmCED~\cite{romdhane2021evaluation} &  MLSVD~\cite{chen2020high} &  DECENT~\cite{chen2023learned} & BOOST~\cite{chen2024boosting} & IRIS~\cite{chen2024implicit} & BALD \\
\hline
1st & 50.71 / 51.51 & 49.30 / 49.38 & 50.95 / 50.96 & 44.76 / 44.79 & 47.15 / 48.54 & 50.19 / 50.65 & \textbf{51.50} / \textbf{51.80} \\ 
2nd & 51.23 / 52.15 & 49.69 / 49.75 & 51.53 / 51.60 & 45.13 / 45.15 & 47.44 / 48.78 & 50.86 / 51.25 & \textbf{52.14} / \textbf{52.51} \\ 
3rd & 50.82 / 51.68 & 49.40 / 49.45 & 51.09 / 51.17 & 44.99 / 45.01 & 46.86 / 48.18 & 50.81 / 50.88 & \textbf{51.66} / \textbf{51.98} \\ 
Average & 50.92 / 51.78  & 49.47 / 49.53 & 51.19 / 51.24 & 44.96 / 44.99 & 47.15 / 48.50 & 50.62 / 50.93 & \textbf{51.77} / \textbf{52.10} \\ 
\hline
\end{tabular}\\
{\footnotesize PSNR (dB) $\uparrow$ is used for evaluation metric, where the best performance of each row is indicated in \textbf{bold}. The metrics for the three noisy scans are 48.9658, 49.3378, and 49.0615, respectively, with an average of 49.1217.}
\label{tab:realphan}
\end{table*}

Validation on physical phantom data leveraged triplicate scans as noisy input, with their averaged sequence serving as the noise-free reference ground truth. Quantitative PSNR analysis (Table~\ref{tab:realphan}, after slash) demonstrated BALD's consistent superiority: individual denoised scans achieved ~3 dB PSNR gains over noisy data, outperforming all baselines across all three replicates. Crucially, this noise suppression directly improved the accuracy of the CEST quantification. Fig.~\ref{fig:realphan_vis} illustrates dopamine contrast (3 ppm) recovery before and after BALD processing across the triplicate scans. In noisy reconstructions, MPF fitting errors manifested as spurious intensity fluctuations that obscured true contrast among the five phantom compartments. After BALD denoising, dopamine maps precisely reconstructed the known concentration hierarchy, with contrast variations perfectly aligned to the reference map. This confirms the unique capacity of BALD to preserve chemically specific exchange signatures while eliminating stochastic acquisition noise, validating its deployment in experimental CEST studies requiring pixel-wise quantification of low-concentration metabolites.

\subsection{\textit{In Vivo} Evaluation}
\begin{figure}
\centering
  \includegraphics[width=1\linewidth]{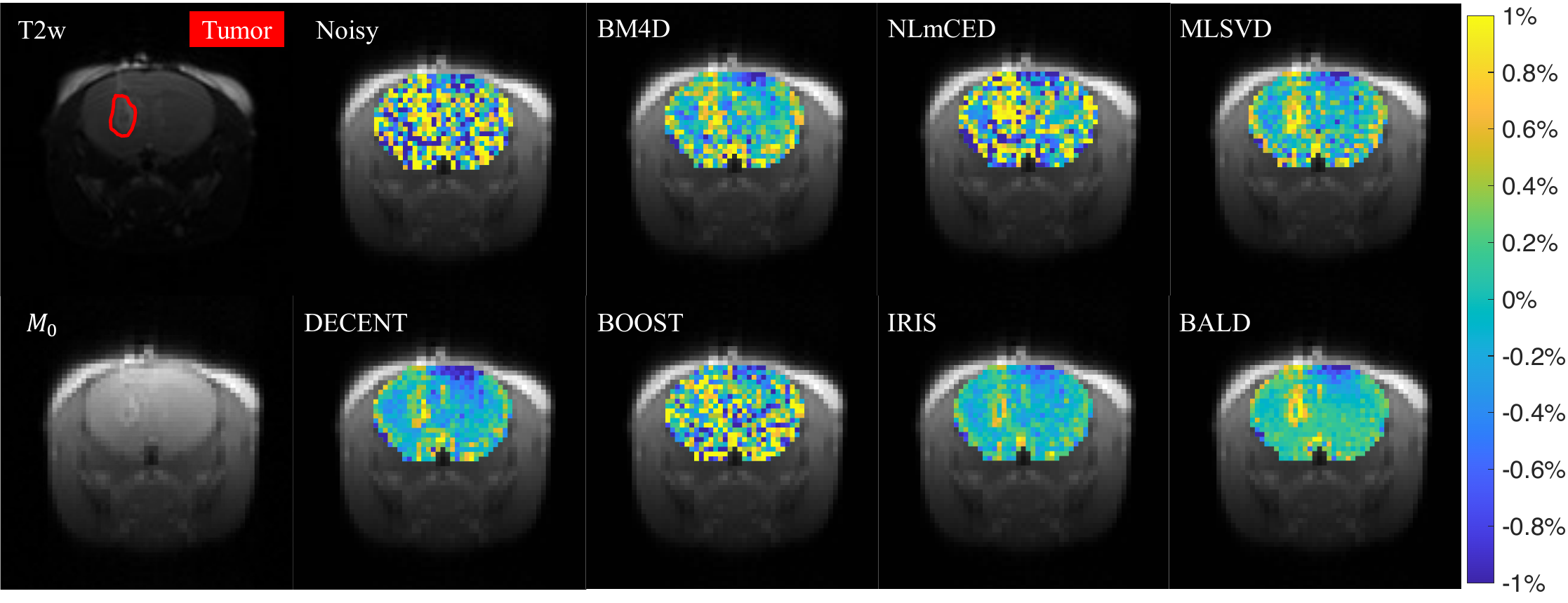}
  \caption{APTw images acquired under $B_1=2\mu T$ and obtained after denoising, with tumor region labeled on the T2-weighted image on the left and the $\textit{M}_0$ image for reference.}
  \label{fig:vivo_mtr}
\end{figure}
\begin{figure*}
\centering
  \includegraphics[width=1\linewidth]{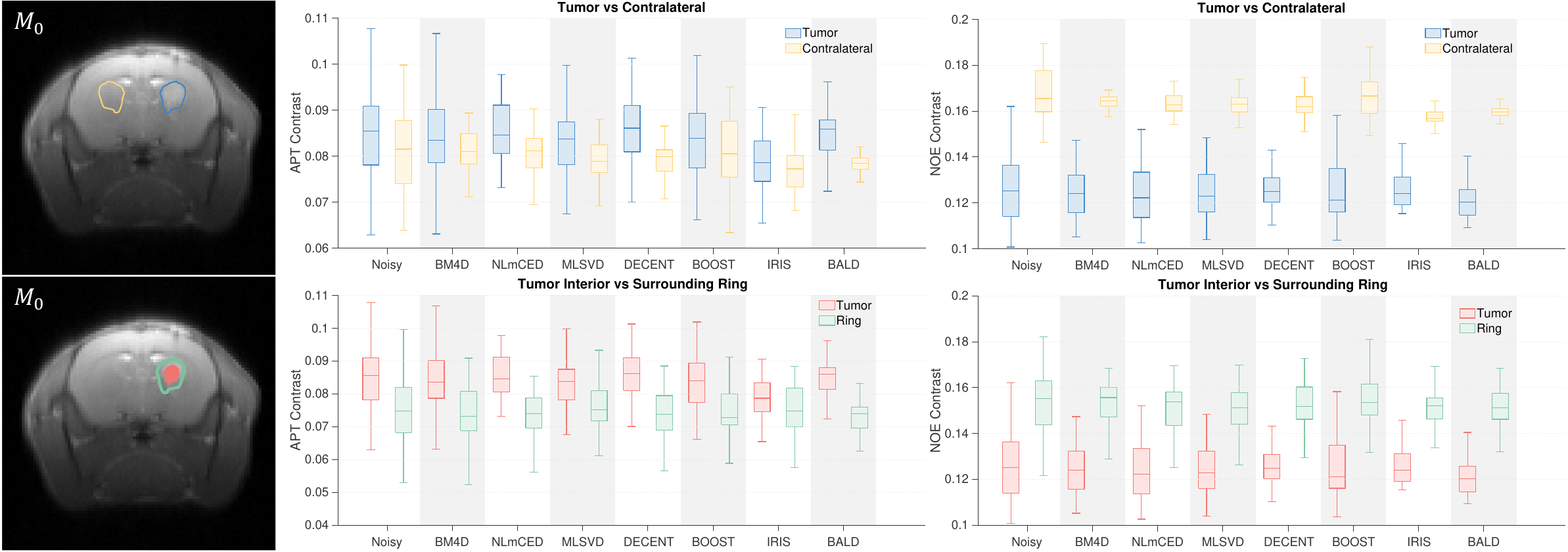}
  \caption{CEST Contrasts intensity distributions in the regions of interest (ROIs), which are labeled on the $\textit{M}_0$ images on the left.}
  \label{fig:vivo_stats}
\end{figure*}
\begin{table}[htbp]
\centering
\caption{T-test about Contrast intensity distributions in the ROIs}
\begin{tabular}{l c c c c}
\hline
        & \multicolumn{2}{c}{Tumor vs Contralateral} & \multicolumn{2}{c}{Tumor vs Ring} \\
         &  APT &  NOE &  APT &  NOE \\ 
\hline
Noisy & 0.3689 & 2.7178e-25 & 1.7720e-05 & 5.8790e-14  \\ 
BM4D~\cite{maggioni2012nonlocal} & 0.3975 & 5.4089e-25 & 1.3531e-06 & 1.5970e-18 \\ 
NLmCED~\cite{romdhane2021evaluation} & 0.1757 & 5.1141e-25  & 5.0940e-08 & 1.4091e-16 \\ 
MLSVD~\cite{chen2020high} & 0.2266 & 1.8917e-27 & 6.4220e-06 & 1.9840e-19  \\ 
DECENT~\cite{chen2023learned} & 3.1646e-04 & 1.6724e-21 & 3.7357e-08 & 5.4545e-09  \\ 
BOOST~\cite{chen2024boosting} & 0.3797 & 3.3612e-28 & 4.7773e-06 & 1.5693e-18  \\ 
IRIS~\cite{chen2024implicit} & 0.05214 & 3.3469e-28  & 1.2722e-05 & 7.5702e-13 \\ 
BALD & \textbf{6.5726e-06} & \textbf{6.1602e-29} & \textbf{4.0410e-13} & \textbf{2.0180e-21} \\
\hline
\end{tabular}\\
{\footnotesize *$p$-value $\downarrow$ is used for evaluation metric, where the best performance is indicated in \textbf{bold}.}
\label{tab:ttest}
\end{table}

\begin{figure}[ht]
\centering
  \includegraphics[width=1\linewidth]{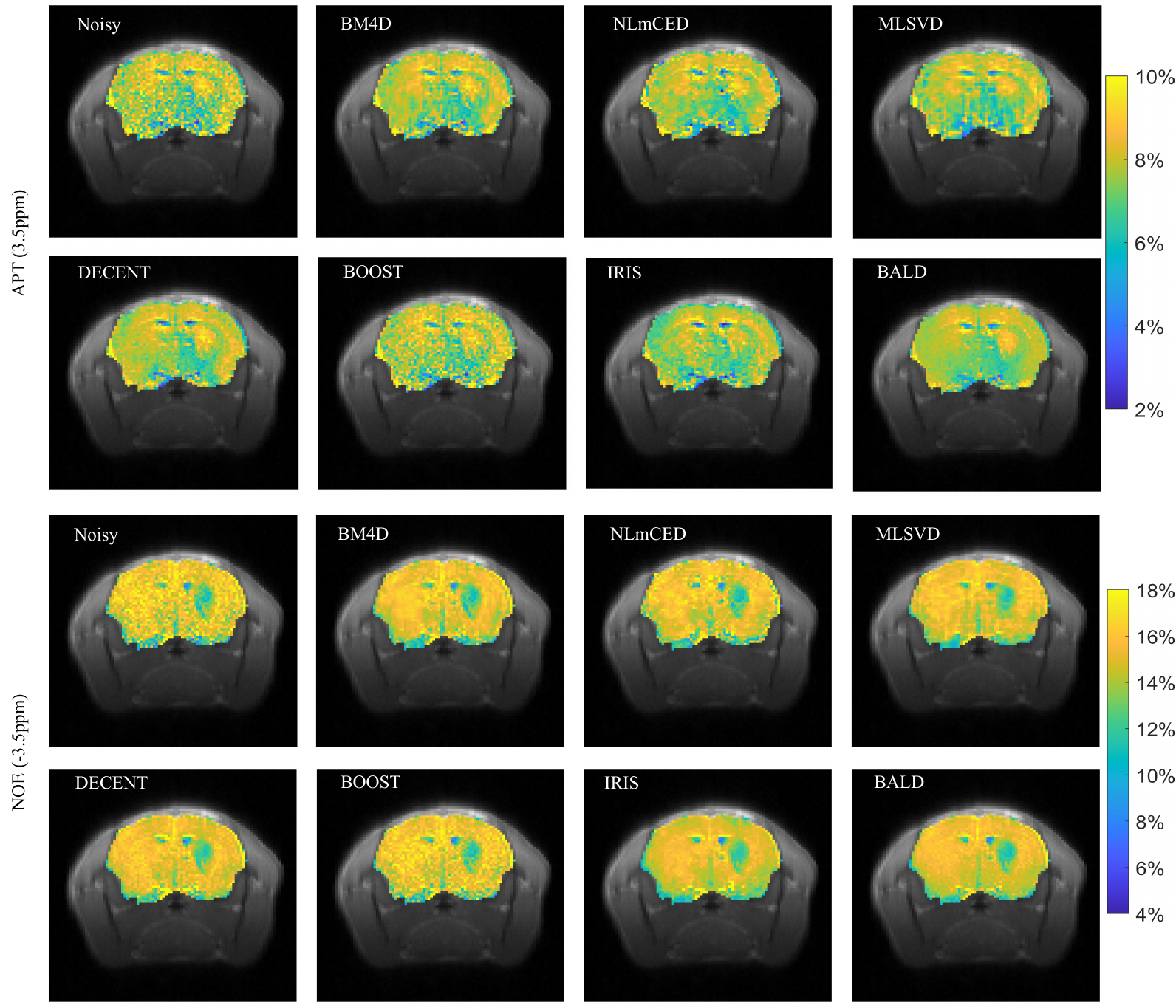}
  \caption{Visualization of MPF-generated CEST contrasts after denoising.}
  \label{fig:vivo_vis}
\end{figure}

\begin{table*}[htbp]
\centering
\small
\caption{Processing Time}
\begin{tabular}{c c c c c c c c}
\hline
Methods     &  BM4D~\cite{maggioni2012nonlocal} &  NLmCED~\cite{romdhane2021evaluation} & MLSVD~\cite{chen2020high} & BOOST~\cite{chen2024boosting} & DECENT~\cite{chen2023learned} & IRIS~\cite{chen2024implicit} & BALD \\
\multirow{2}{*}{Time (s)} & \multirow{2}{*}{53.3339} & \multirow{2}{*}{54.3899} & \multirow{2}{*}{57.2031} & \multirow{2}{*}{40.2289} & 2.7163 & \multirow{2}{*}{57.7468} & \multirow{2}{*}{35.1909}  \\
                        &&&&& (4 days training) & & \\
\hline
\end{tabular}
\label{tab:time}
\end{table*}

The clinical relevance of BALD was validated through the \textit{in vivo} brain tumor mouse model, where CEST contrast mapping critically differentiates pathological features.

In particular, we analyzed the contrast established by $\text{MTR}_{\text{asym}}$, which is inherently sensitive to noise due to its calculation relying on only two frequency offsets. According to the consensus at 3T in brain tumors~\cite{zhou2022review}, APTw imaging at $B_1 = 2\mu T$ serves as an important clinical benchmark. Therefore, this experiment was designed to validate and compare the effectiveness of upstream denoising methods. As shown in Fig.~4\ref{fig:vivo_mtr}, with reference to the left-side T2w and $\textit{M}_0$ images indicating tumor location and morphology, the APTw intensity computed after BALD denoising highly correlates with the tumor region. Moreover, the internal hollow tissue structure within the tumor region at this slice is clearly visible in the BALD. Additionally, the APTw intensity in the contralateral normal-appearing white matter (CNAWM) remains around $0\%$, which is highly consistent with findings in the literature~\cite{zhou2022review}. In contrast, other denoising methods resulted in unclear APTw contrast due to suboptimal denoising performance.

Tumors also exhibit hyperintense APT signals and hypointense NOE signals relative to CNAWM, reflecting molecular alterations characteristic of malignancy~\cite{huang2022molecular}. These signature patterns provide a biological ground truth to evaluate whether denoising preserves essential diagnostic information while suppressing noise.

Fig.~\ref{fig:vivo_stats} illustrates the analytical framework: the left column demarcates key regions-of-interest (ROIs) on $\textit{M}_0$ anatomy, tumor (blue), CNAWM (yellow), tumor interior (red) and tumor surrounding ring (green), while the right panel compares post-denoising APT/NOE contrast distributions via boxplots. Critically, noisy or insufficiently denoised contrasts (e.g., BM4D, BOOST) exhibited inflated variance within homogeneous tissues (Fig.~\ref{fig:vivo_vis}), contradicting biological consistency where uniform tissue biochemistry should yield compact signal distributions. BALD uniquely achieved optimally condensed distributions in both APT and NOE maps (Fig.~\ref{fig:vivo_stats}), with tumor-versus-CNAWM contrasts aligning precisely with expected pathological signatures: elevated APT and reduced NOE in neoplastic regions.

The lower boxplots quantify tumor contrast, revealing that BALD maximized contrast separation between lesions and background parenchyma. This is visually corroborated in Fig.~\ref{fig:vivo_vis}, where BALD-generated maps display sharpened tumor boundaries and enhanced textural differentiation compared to blurred outputs from other denoising baselines. Statistical validation in Table~\ref{tab:ttest} demonstrates BALD’s superior preservation of pathological discriminability: it achieved the lowest $p$-values in all four significance tests, tumor vs. CNAWM (APT/NOE) and tumor interior vs. ring (APT/NOE), confirming its unparalleled ability to amplify biologically meaningful contrasts. Notably, among all methods, only DECENT and BALD maintained a diagnostically confident APT contrast in the tumor characteristic, though DECENT incurred a spatially non-uniform APT intensity distribution of the tumor.

This \textit{in vivo} analysis confirms that the adaptive stabilization and local orthogonal filtering of BALD synergistically eliminate noise without erasing subtle pathological gradients, establishing it as an enabling tool for quantitative CEST-based disease characterization.

In addition, we compared the computational efficiency as in Table~\ref{tab:time}. Although DECENT processes the same CEST sequence with the shortest time, it requires days of training, with training dataset simulation time excluded, while BALD achieves the best efficiency among the unsupervised baselines.

\subsection{Noise Modeling and Self-ablations}
\label{sec:noise_modeling}

\begin{figure}
\centering
    \subfigure[Real Phantom]{
{\includegraphics[width=0.31\linewidth]{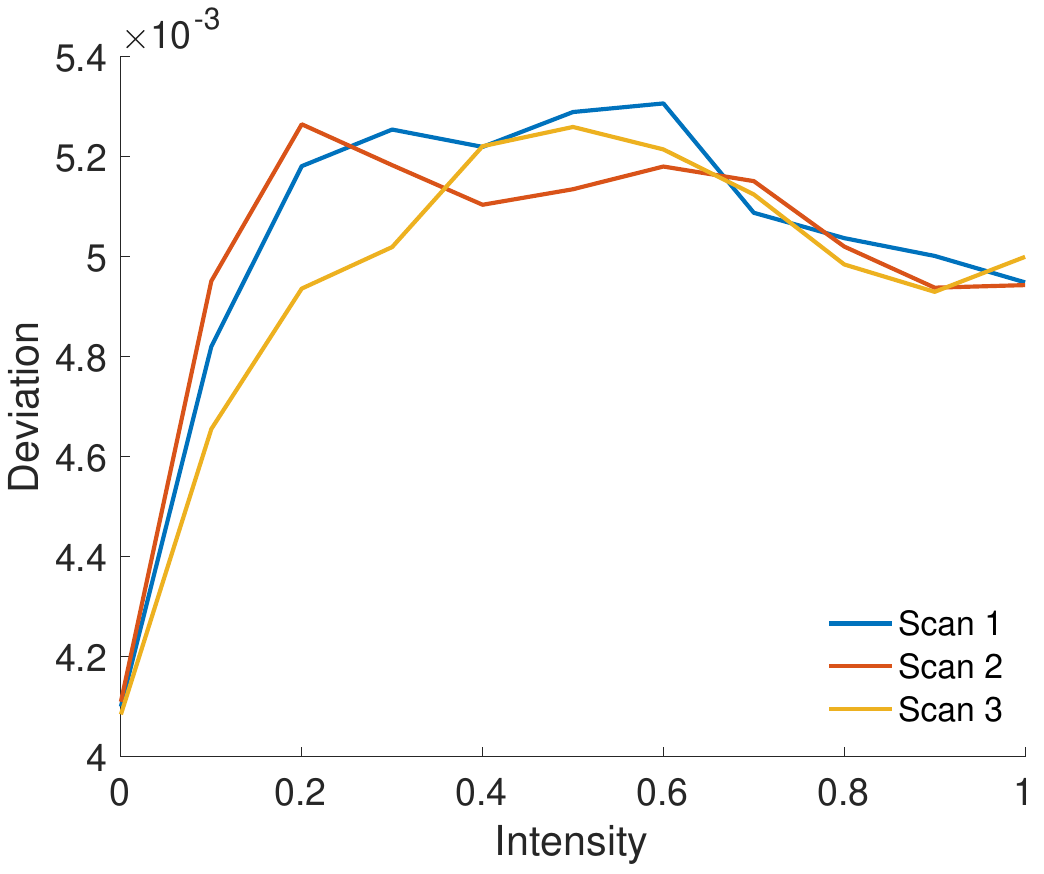}}}
   \subfigure[Synthetic Phantom]{
{\includegraphics[width=0.31\linewidth]{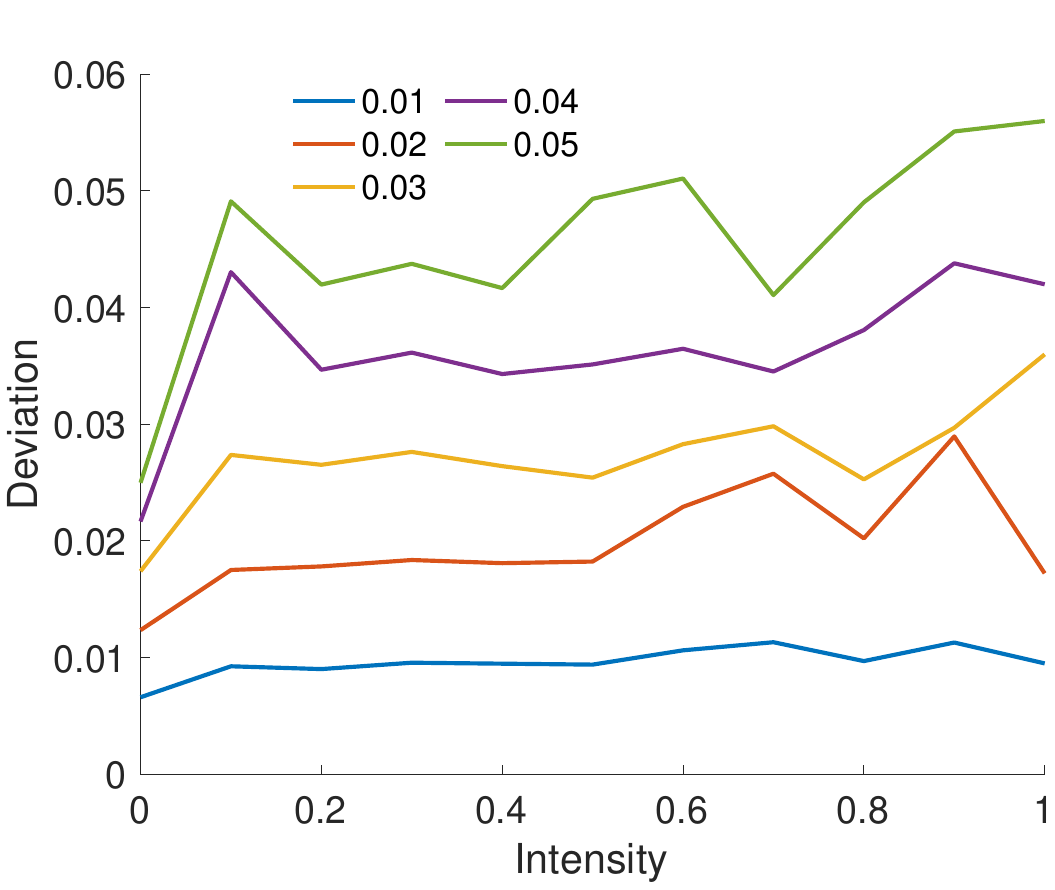}}}
    \subfigure[In-vivo Subjects]{
{\includegraphics[width=0.31\linewidth]{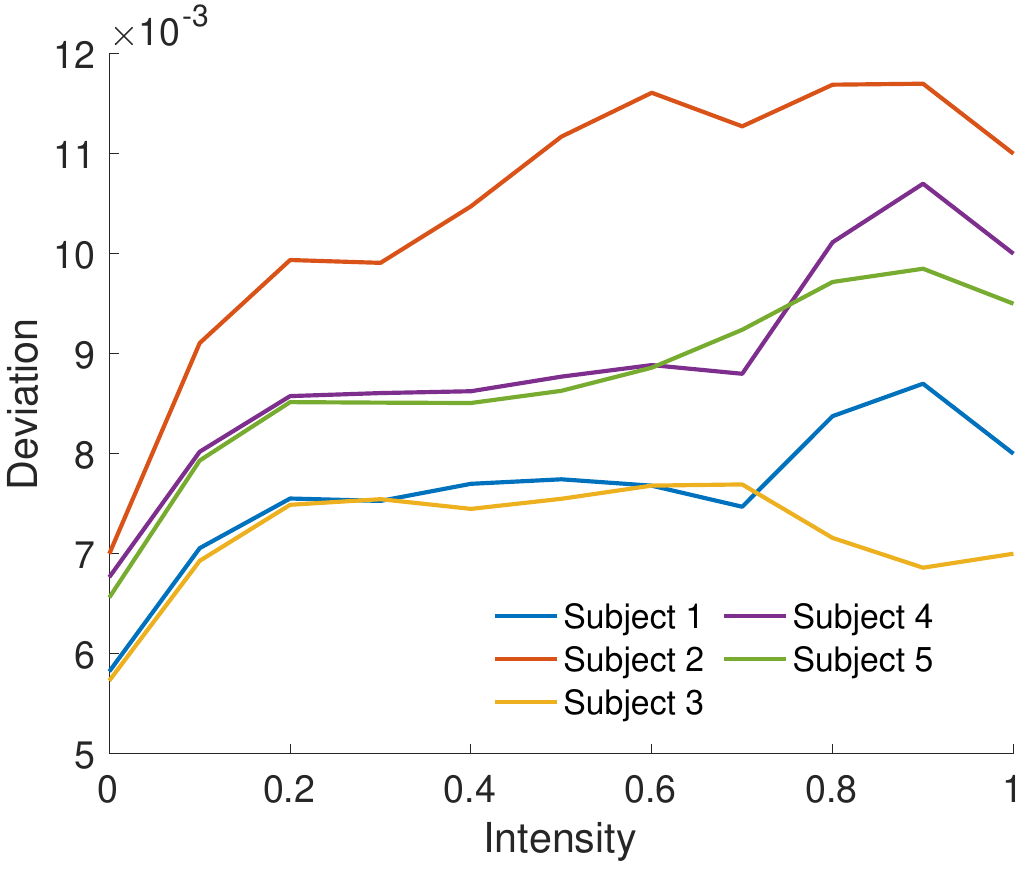}}}
    \subfigure[Noise Estimation from Given Noise Models]{
{\includegraphics[width=0.7\linewidth]{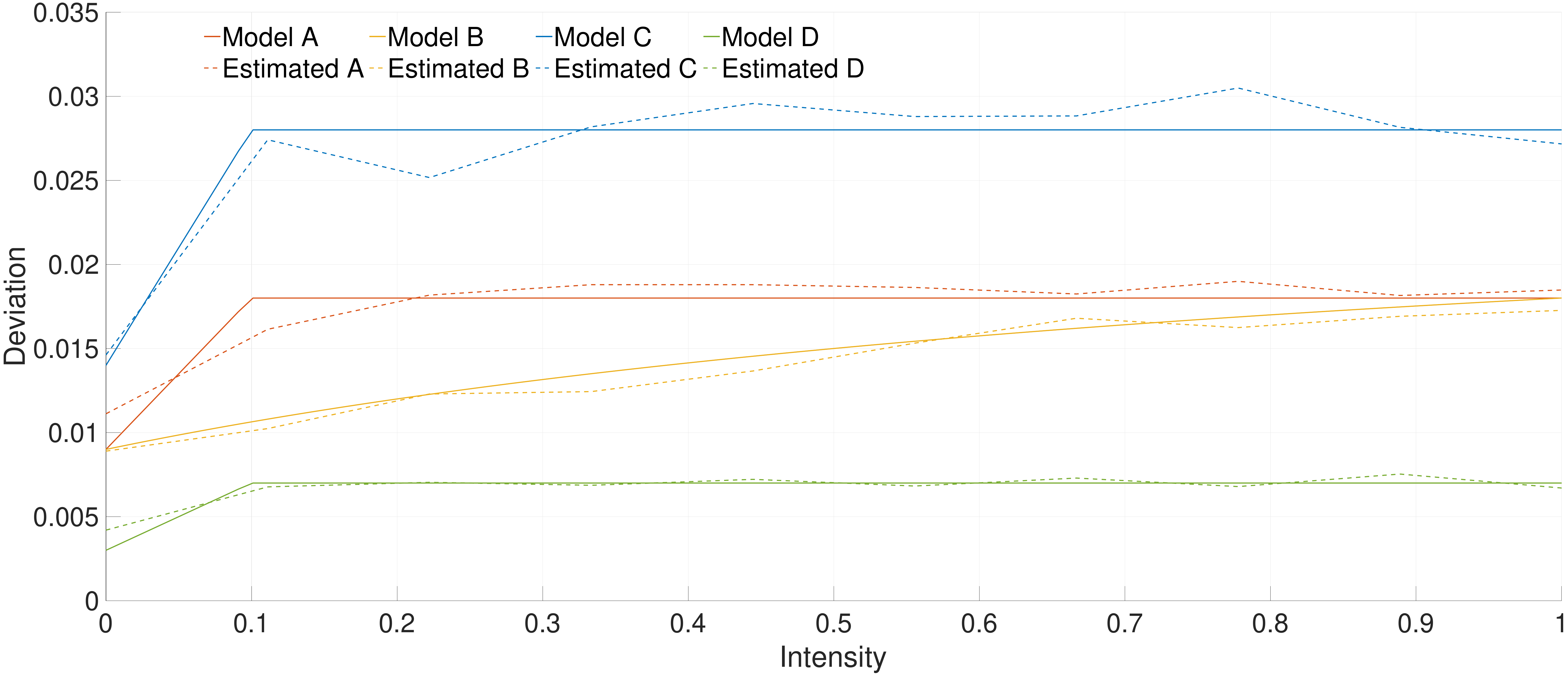}}}
    \caption{Evaluation on the accuracy of noise model estimations. According to the noise models estimated from (a) real phantoms, (b) synthetic phantoms with multi-level Riccian noise, and (c) \textit{in vivo} subjects, we empirically constructed four similar and ideal noise models to simulate noisy phantoms. The dashed plots in (d) demonstrate our precise noise model estimations.}
    \label{fig:noisemodel}
\end{figure}

\begin{table}[htbp]
\centering
\caption{Ground Truth Noise Model vs Model Used to Denoise}
\label{tab:modeling}
\begin{tabular}{c|c c c c}
\hline
\bfseries GT\textbackslash Used &  A &  B &  C &  D \\ 
\hline
A & \textbf{24.9654} & 24.9130 & 24.6988 & 24.7670 \\ 
B & 25.0585 & \textbf{25.0815} & 25.0793 & 24.9406 \\ 
C & 22.6455 & 22.5622 & \textbf{22.6845} & 22.3855 \\ 
D & 28.0868 & 28.0811 & 28.0935 & \textbf{28.1587} \\ 
\hline
\end{tabular}\\
{\footnotesize *PSNR (dB) $\uparrow$ is used for evaluation metric, where the best performance is indicated in \textbf{bold}.}
\end{table}

The precision of noise modeling emerged as a critical determinant of denoising performance across all datasets. Empirically derived noise models for real phantoms, synthetic phantoms, and \textit{in vivo} data (Fig.~\ref{fig:noisemodel} (a)-(c)) consistently revealed a nonlinear deviation profile: noise variance escalated sharply at low intensities but plateaued in mid-to-high signal regimes, confirming the signal-dependent noise structure inherent to CEST MRI. To validate modeling accuracy, four idealized analytical models $g(u)$ (Fig.~\ref{fig:noisemodel} (d), solid lines) were applied to synthetic data via Eq.~\ref{eq:noise}, generating noisy datasets (A)-(D). Remarkably, when our methodology reconstructed noise models (Section \ref{sec:avst}) from these datasets, the estimates (dashed lines) converged almost perfectly to ground truth (Fig.~\ref{fig:noisemodel} (d)), demonstrating the method's capability to capture ground-truth noise characteristics. Subsequent denoising of (A)-(D) using deliberately mismatched models in BALD (Table~\ref{tab:modeling}) proved unequivocally that optimal performance requires precise noise-deviance alignment. This establishes that: (1) denoising efficacy fundamentally depends on noise model accuracy, and (2) the adaptive estimation of BALD reliably captures true noise physics from raw CEST sequences.

Further ablation studies dissected the architectural synergy of BALD. When disabling AVST (Table~\ref{tab:realphan}, before slash), standalone Local SVD filtering still outperformed conventional methods, attributable to its structure-preserving orthogonal decomposition that mitigates spectral leakage without variance stabilization. However, integrating AVST universally elevated all methods' performance (Table~\ref{tab:realphan}, after slash), including VST-equipped baselines (e.g., BM4D, BOOST) retrofitted with our adaptive transform. The increment of PSNR confirms AVST's unique capacity to condition CEST-specific noise into stationary Gaussian domains where local filters operate optimally. Crucially, the full pipeline of BALD synergized these advantages multiplicatively, delivering gains unattainable by either component alone, validating its co-design philosophy for improving the sensitivity of CEST imaging.

\subsection{Performance Refinement for Pre-trained Models}
\begin{table}[htbp]
\centering
\caption{Pre-trained models performance refinement with AVST}
\begin{tabular}{c c c c}
\hline
                 &  DnCNN~\cite{zhang2017dncnn} &  FFDNet~\cite{zhang2018ffdnet} & DRUNet~\cite{zhang2021plug} \\
\hline
Original Model & 45.9769 & 47.5146 & 49.2851 \\
AVST & 47.2731 & 49.1958 & 50.4443 \\
\hline
\end{tabular}\\
\label{tab:pretrain}
{\footnotesize *PSNR (dB) $\uparrow$ is used for evaluation metric. PSNR of the noisy data is 49.1217 on average.}
\end{table}

AVST is not merely a component of BALD, but a standalone meta-algorithm that empowers any Gaussian-optimized denoiser to conquer the challenges of CEST MRI. We demonstrate how AVST enables unprecedented generalization of natural-image-trained deep models to CEST MRI. As validated in Table~\ref{tab:pretrain}, applying AVST as a preprocessing step universally elevated the performance of denoisers (e.g., DnCNN~\cite{zhang2017dncnn}, FFDNet~\cite{zhang2018ffdnet}, DRUNet~\cite{zhang2021plug}) originally optimized for additive white Gaussian noise in natural images. When confronted with raw CEST sequences exhibiting complex signal-dependent noise, these models suffered substantial performance degradation (e.g., DnCNN). However, feeding them AVST-transformed data, where noise statistics become stationary and Gaussian, achieves consistent 0.6–0.7 dB average PSNR gains over native CEST processing. This refinement occurs because AVST synthetically converts the pathological noise distribution of CEST data into the canonical Gaussian structure recognized by pre-trained networks, effectively tricking them into applying their learned priors optimally. Critically, this approach bypasses the prohibitive costs of medical-specific retraining while leveraging massive natural image corpora, establishing AVST as a universal adapter that unlocks cross-domain knowledge transfer for computational imaging and enabling immediate deployment of cutting-edge photographic AI models in clinical CEST pipelines without architecture modifications or medical data collection.

%% file: Conlusion.tex
\section{Conclusion}
This study introduces the Blind Adaptive Local Denoising (BALD) framework, a paradigm-shifting solution for CEST MRI denoising that fundamentally rethinks noise management through two synergistic innovations. First, our proposed Local SVD algorithm achieves unprecedented noise suppression via a dual-stage orthogonal projection architecture: hard thresholding eliminates stochastic noise components while Wiener filtering performs signal-adaptive refinement, preserving critical spectral-spatial correlations unique to CEST data. Second, the Adaptive Variance Stabilization Transform (AVST) establishes a new standard for noise modeling, dynamically converting signal-dependent noise into stationary Gaussian domains through curvature-aware deviation estimation. Rigorous validation across synthetic phantoms, real phantoms, and \textit{in vivo} tumor models confirms BALD's supremacy, outperforming state-of-the-art methods while enhancing downstream CEST contrast in pathological discrimination and ROI identification. Crucially, the transformative impact of AVST extends beyond BALD: when integrated into conventional denoisers or natural-image-trained deep networks, it bridges the domain gap to CEST imaging, boosting performance without architectural modifications. This dual capability, exceptional standalone denoising, and universal noise conditioning position BALD as both a performance benchmark and an enabling technology. By ensuring accurate quantification of macromolecules and metabolites, even in low SNR conditions, our framework unlocks new frontiers in CEST-based biochemical profiling and disease mechanism research, from CEST contrast mapping to early tumor characterization.